\documentclass[letterpaper, 10 pt, journal, twoside]{IEEEtran}

\usepackage{times}
\usepackage[pdftex]{graphicx}
\usepackage{amsmath,amssymb,amsopn,amstext,amsfonts}
\usepackage{cancel}
\usepackage[space]{cite}
\usepackage{pdfsync}
\usepackage{balance}
\usepackage{color}
\usepackage{mathtools}
\usepackage{algpseudocode}
\usepackage[ruled,vlined]{algorithm2e}
\usepackage{bm}

\usepackage{diagbox}
\usepackage{float}
\usepackage{epstopdf}
\usepackage{pifont}
\usepackage{amsmath}
\usepackage{multirow}
\usepackage{url}
\usepackage{verbatim}
\usepackage[linkcolor=black,citecolor=black,urlcolor=black,colorlinks=true]{hyperref}
\usepackage{booktabs}
\usepackage{graphicx}
\usepackage{subcaption}
\usepackage{threeparttable}  
\usepackage{tikz}
\newcommand*{\circled}[1]{\lower.7ex\hbox{\tikz\draw (0pt, 0pt)%
		circle (.5em) node {\makebox[1em][c]{\small #1}};}}

\graphicspath{{./figures/}}
\DeclareGraphicsExtensions{.png,.jpg,.eps,.pdf}
\IEEEoverridecommandlockouts

\newcommand{\tp}{^{\mathrm{T}}}

\newcommand{\deletefig}[1]{{\bgroup\markoverwith{\textcolor{red}{\rule[2.5ex]{2pt}{2.0pt}}}\ULon{#1}}}
\newcommand{\add}[1]{{\color{black}{#1}}}

\newcommand{\delete}[1]{}

\author{Qianhao Wang $^{\dag \,}$\textsuperscript{1,2}, 
	Botao He $^{\dag \,}$\textsuperscript{2,3}, 
	Zhiren Xun \textsuperscript{1,2},
	Chao Xu \textsuperscript{1,2},
	and Fei Gao\textsuperscript{1,2}
	\thanks{Manuscript received: September, 9, 2021; Revised: December, 21, 2021;
	Accepted: January, 23, 2022.
	This paper was recommended for publication by Editor Pauline Pounds upon evaluation of the Associate Editor and Reviewers' comments. This work was supported by National Natural Science Foundation of
	China under Grant 62003299 and Grant 62088101. (Corresponding author: Fei Gao, Chao Xu.)}
	\thanks{\textbf{${\dag}$ Equal contribution.}}        
	\thanks{1 State Key Laboratory of Industrial Control Technology, Institute of Cyber-Systems and Control, Zhejiang University, Hangzhou, 310027, China.} 
	\thanks{2 Huzhou Institute of Zhejiang University, Huzhou, 313000, China.}
	\thanks{3 School of Automation, Nanjing Institute of Technology, Nanjing, 211112, China.} 
	\thanks{Email:{\tt\small \{qhwangaa, fgaoaa\}@zju.edu.cn}}
	\thanks{Digital Object Identifier (DOI): see top of this page.}
}

\title{GPA-Teleoperation: Gaze Enhanced Perception-aware Safe Assistive Aerial~Teleoperation}

\begin{document}
%
    
    \maketitle
    
	\begin{abstract}
		Gaze is an intuitive and direct way to represent the intentions of an individual.
		However, when it comes to assistive aerial teleoperation which aims to perform operators' intention, rare attention has been paid to gaze.
		Existing methods obtain intention directly from the remote controller (RC) input, which is unfriendly to non-professional operators, as the experimental results show in Sec. \ref{sec:EvaTopo}.
		Further, most teleoperation works do not consider environment perception which is vital to guarantee safety.
		In this paper, we present GPA-Teleoperation, a gaze enhanced perception-aware assistive teleoperation framework, which addresses the above issues systematically.
		We capture the intention utilizing gaze information, and generate a topological path matching it.
		Then we refine the path into a safe and feasible trajectory which simultaneously enhances the perception awareness to the environment which the operator is interested in.
		Additionally, the proposed method is integrated into a customized quadrotor system.
		Extensive challenging indoor and outdoor real-world experiments and benchmark comparisons verify that the proposed system is reliable, robust and applicable to even unskilled users.
		We will release the source code\footnote{https://github.com/ZJU-FAST-Lab/GPA-Teleoperation} of our system to benefit related researches.
	\end{abstract}

    
\section{Introduction}
\label{sec:Introduction}

\IEEEPARstart{W}{ith} the rapid development of aerial autonomy, UAV assistive teleoperation gradually shows huge potentials to reduce the difficulty of operations and raise the safety of missions.
Assistive teleoperation, which aims to accomplish humans' intention by machines, puts forward twofold requirements
. 
Firstly, it should capture the intention of operators precisely and timely. 
Secondly, it should be accompanied by a proper acting strategy under the promise of safety. 

Gaze fixations are strongly correlated to the operator’s intention, which has been observed in various fields, including car driving \cite{land1994we}, gaming \cite{belkacem2015real}, and drone racing \cite{pfeiffer2021human}. 
Naturally, gaze has an non-negligible potential for representing intentions.
However, in the literature on aerial teleoperation, few works try to utilize the gaze of humans.
Instead, most existing works \cite{yang2020assisted, israelsen2014automatic, yangintention}
directly read the intention input from 
RC, which may not be able to reflect the true intention of naive operators. This is because using RC requires the operator to simultaneously operate at least four channels to control drone's direction and speed, which is hard for those who do not have long-term practices.
These unstable intentions are particularly obvious when the drone is operated by an unskilled man in an unfamiliar environment and may prevent the assistive flight system from being truly accepted by consumers.
In this paper, inspired by the work \cite{pfeiffer2021human} which finds the high correlation between gaze and intention, we propose an aerial assistive teleportation system, which stably captures the human intention by using the gaze information, and react to this intention in a user-friendly way. 

\begin{figure}[t]
	\vspace{0.0cm}
	\centering
	\includegraphics[width=0.9\linewidth]{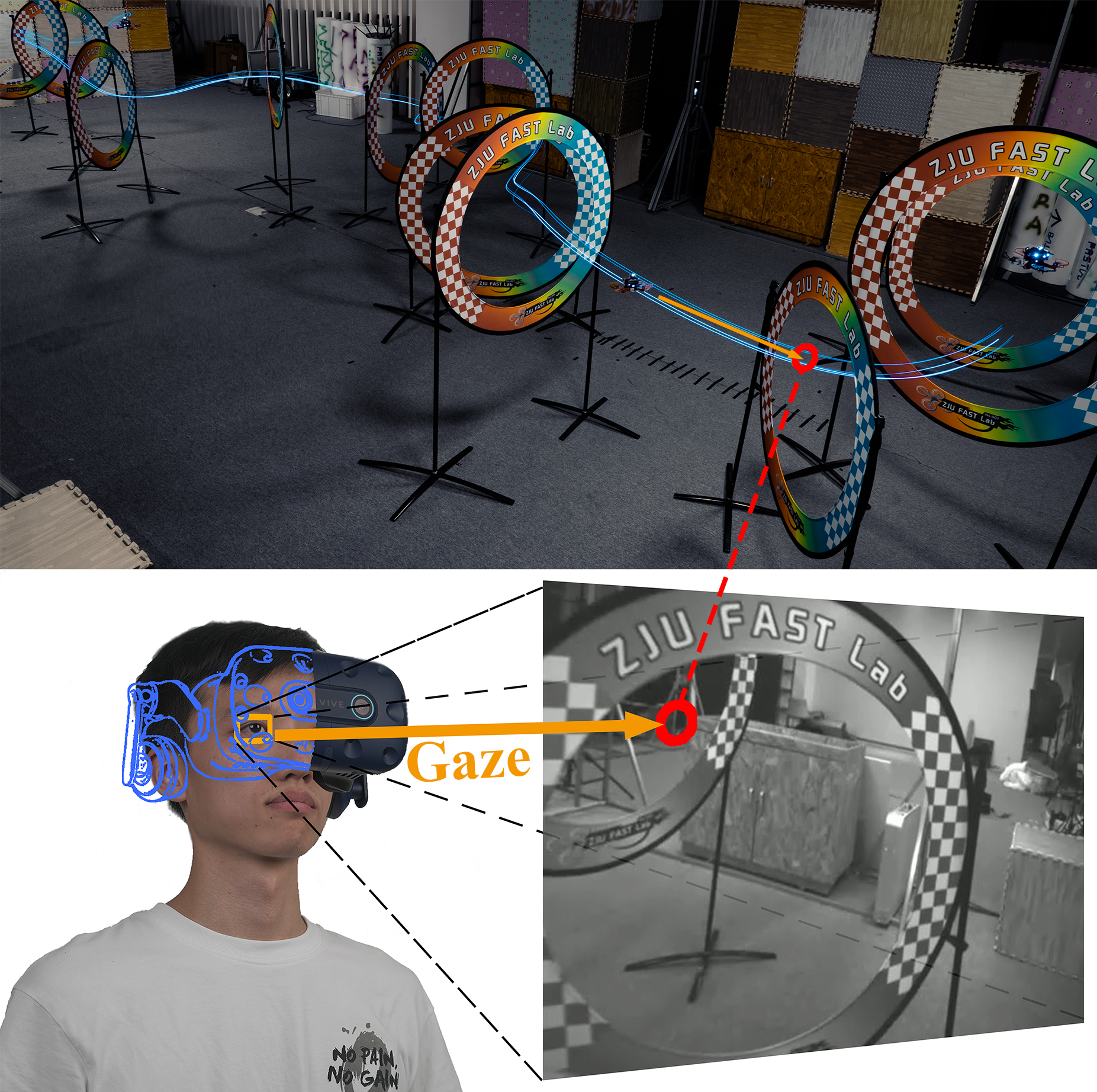}
	\vspace{-0.1cm}
	\captionsetup{font={small}}
	\caption{
		Illustration of the real-world drone-racing experiment, which will be detailed in Sec. \ref{sec:EvaTopo}.
		\textbf{Top}: the composite long exposure trajectory of quadrotor flying with our assistive teleoperation system.
		\textbf{Bottom}: the operator with an eye tracker and the image he receives now is from the first view of the quadrotor in the yellow box.
		Red circles in the top and bottom figure represent gaze data in three dimensions and the first view respectively.
	}
	\label{fig:cover}
	\vspace{-1.0cm}
\end{figure}

In addition to intention inference, safety is another pivotal element for aerial assistive teleoperation.
Since the intention may originate from an unskilled operator, increasing the safety margin in occluded and complex environments is vital.
Therefore, a robust flying assistance system should not only generate safe motions using operator's intention, but also assist operators to do timely decisions with environmental awareness. 
Recently, many works \cite{heiden2017planning}, \cite{zhou2020raptor}, including our previous research \cite{wang2021visibility} start to consider perception awareness by planning the drone to observe and avoid unknown obstacles actively.
Nevertheless, these works cannot be directly applied to assistive aerial teleoperation (details in Sec. \ref{sec:RelatedWorks::PerceptionAware}). 
To address this issue, we propose a novel perception-aware planner by improving our previous work \cite{wang2021visibility} to enhance the visibility of where the operator's intention refers. 

Furthermore, keeping the speed of the drone on par with the operator's expectation is also essential in a user-friendly teleopeartion system. This is because frequent unexpected speed changes will make the operator feel like they are losing effective control of the drone. To address this problem, we upgrade our trajectory planner to maintain consistent with the operator's expected speed as much as possible.

In this paper, we propose a \textbf{G}aze enhanced \textbf{P}erception-aware safe \textbf{A}ssistive aerial teleoperation system called \textbf{GPA-Teleoperation}. 
Based on eye movements and RC input, we plan a topological path that matches operator's intention. 
Derived from this path, we further propose the back-end optimizer which refines the path into a 
trajectory which simultaneously ensures safety and increases the perception awareness to environment of interest.
Additionally, we integrate our proposed system into a customized quadrotor.

Additionally, we compare our system with existing consumer class assistive aerial teleoperation by several operators. 
Extensive real-world experiments show our method performs efficiently and safely.
The contributions of this paper are:
\begin{itemize}
	\item [1)] 
	A novel intention capture method utilizing gaze fixation, which generates a guiding topological path for trajectory optimization. Precise intention obtained from this method makes the system more friendly to operators.
	\item [2)]
	A perception-aware trajectory optimization method,
	which simultaneously optimizes the visibility to environments
	where the operator focuses, and maintains
	the speed that the operator desires. It promotes the
	safety and user experience of the proposed system.
	\item [3)]
	A complete gaze enhanced assistive teleoperation system with code open source.
	Extensive real-world tests validate that our method is practical and effective.
	
\end{itemize}

\section{Related Works}
\label{sec:RelatedWorks}

	\subsection{Gaze for Intention Inference}
	There have been many works, mainly for car driving scenarios, studying gaze-based intention inference. 
	With Land et al. \cite{land1994we} first illustrate the high relevance between gaze and future trajectory, several works repeatedly confirm this both in real-world \cite{boer1996tangent}, \cite{lappi2013beyond} and simulation \cite{negi2019differences}. 
	For UAVs, the phenomenon has also been observed. 
	In \cite{hubenova2020usage}, the authors notice that eye movement patterns are specific in different flight stages like take-off and landing. 
	In \cite{pfeiffer2021human}, Pfeiffer et al. find that gaze fixations are highly correlated with operator's control commands, which means eye movements can well reflect the operator's intention. 
	These works provide solid evidence that humans tend to use specific gaze patterns to indicate their intents. 
	However, none of these works integrates human gaze into a closed control loop. 
	While several works in visual servoing \cite{valenti2012you}, \cite{shi2019application} indicate that using gaze to control robots for specific tasks are intuitive and operators do not need any long-term professional training.

	\subsection{Assistive Aerial Teleoperation}
	Several previous works \cite{nieuwenhuisen2013multimodal},  \cite{odelga2016obstacle} formulate the assistive teleoperation of UAVs as a local control problem to avoid collisions. 
	However, these works do not take operator's intention into account.
	In recent years, some works have been proposed to incorporate operator's intention into assistive teleoperation.
	Yang et al. \cite{yang2020assisted} handle the intention of an operator by modeling it as an indicative direction and follow it by generating trajectories from a set of motion primitives.
	However, this sampling-based method is time-consuming and cannot guarantee a promising trajectory due to the inherent discretization error.
	Recently, Yang et al. \cite{yangintention} propose a hierarchical teleoperation framework considering global intentions.
	They capture intention of the operator to generate a global path and plan local safe trajectories while following it.
	The above works capture intentions directly from the inputs of a RC, which is unfriendly to operators without long-term training in challenging environments. 
	However, by capturing intentions precisely and stably with gaze information, our system is applicable to even naive users.
	
	\subsection{Perception-aware Navigation in Unknown Environments}
	\label{sec:RelatedWorks::PerceptionAware}
	For UAVs, generating safe flight trajectories is the key to developing a navigation system.
	Richter et al. \cite{richter2016learning} propose a learning-based method to plan for better visibility to unknown areas. 
	However, the huge amount of data and training prevents it from being generalized to complex 3D environments.
	Zhou et al. \cite{zhou2020raptor} propose a perception-aware planning strategy. 
	They use an iterative task-specific trajectory refinement method to achieve risk awareness. 
	This method guarantees sufficient safe reaction distance for suddenly appeared obstacles. 
	In this paper, we extend our previous work \cite{wang2021visibility} to equip our assistive teleoperation system with environmental awareness.
	Our method successfully increases confidence and safety when operators teleoperate the drone in complex environments.

\begin{figure*}[ht]
	\vspace{0.0cm}
	\centering
	\includegraphics[width=\textwidth]{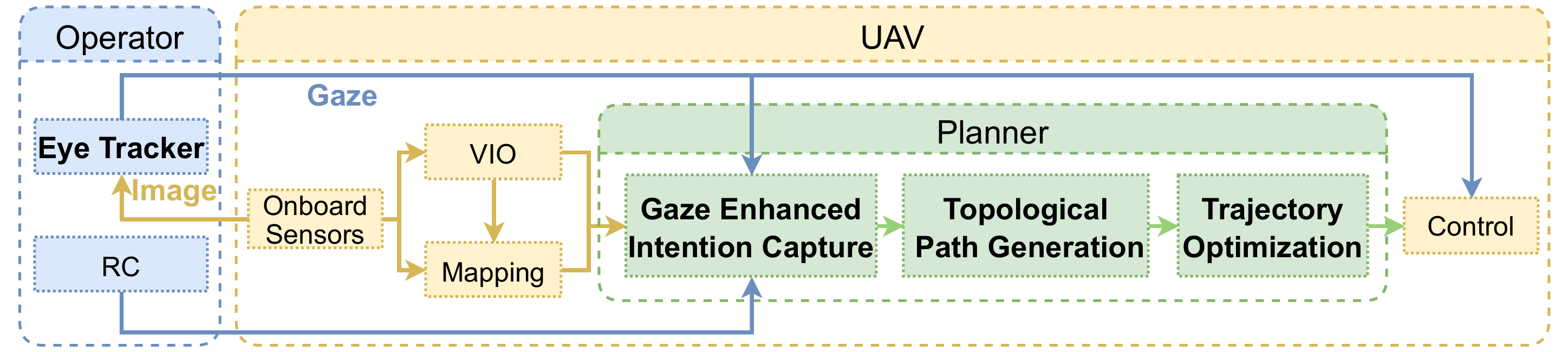}
	\vspace{-0.6cm}
	\captionsetup{font={small}}
	\caption{
		A diagram of our teleoperation system architecture.
	}
	\label{fig:system}
	\vspace{-0.5cm}
\end{figure*}

\section{System Overview}
\label{sec:Overview}

The problem statement of our work is: capturing operator's direction intention from gaze and speed intention from RC input, our system generates a safe trajectory that conforms to the intention for the UAV to execute.
The pipeline of this work is illustrated in Fig. \ref{fig:system}.
Our teleoperation system is divided into two subsystems, operator interaction subsystem and UAV subsystem.

For the operator interaction subsystem, eye tracker and RC are used  respectively to obtain operator's expected 
direction and speed.
Since RC is only used for speed control, we only need a single channel input from it, which is similar to the accelerator of a vehicle.
Eye tracker receives the first view image from the UAV's onboard sensor.

For the UAV subsystem, state estimation, mapping, planning and control modules are all working onboard. 
Based on 
expected direction and speed, the planner generates a local target and the operator's desired topological path, which will be detailed in Sec. \ref{sec::TopoPathGen}.
Then by refining this topological path, the planner optimizes a perception-aware trajectory that 
makes the UAV keep at operator’s desired speed, as presented in Sec. \ref{sec::TrajectoryOPtimization}.
Additionally, to allow operator to view the intended topology better, we control the yaw of UAV to place the topology around the center of the view.

\begin{algorithm}
	\caption{Topological Path Generation}
	\label{alg:TopoPathGen}
	let $Q$ be a queue; 
	label $g_{D}$ as explored\;
	$Q.enqueue(g_{D})$\;
	\While{$Q$ is not empty}
	{
		$v := Q.dequeue()$\;
		\If{\delete{$v.depth \in 0.3~5m$} \add{$v.depth$ in perception range} \textbf{and} $v$ is unclustered}{
			$c_{temp} := \mathbf{DBSCAN}(v, D)$\;
			label $c_{temp}$ as clustered\;
			\eIf{$g_{D} \in \mathbf{BoundingBox}(c_{temp})$}{
				$\mathcal{C}.enqueue(c_{temp})$;
			} {
				\eIf{$c^{1}_{m}$ is empty}{
					$c^{1}_{m} = (c_{temp})$\;
					$\mathcal{C}.enqueue(c^{1}_{m})$\;
				}{
					$\mathcal{M} = \mathbf{Merge}(c^{1}_{m}, c_{temp})$\;
					\If{$g_{D} \in \mathbf{BoundingBox}(M)$}{
						$c^{2}_{m} = c_{temp}$\;
						$\mathcal{C}.enqueue(c^{2}_{m})$\;
						\textbf{Break}\;
					}
				}
			}
		}
		$Q.enqueue(\mathbf{VisitNeighbours}(v))$; \tcp{BFS}
		
	}
	\Return{$\mathcal{C}$}\;
\end{algorithm}

\section{Gaze Enhanced Intention Guiding Topological Path Generation}
\label{sec::TopoPathGen}

\begin{figure}[t]
	\centering
	\includegraphics[width=0.85\linewidth]{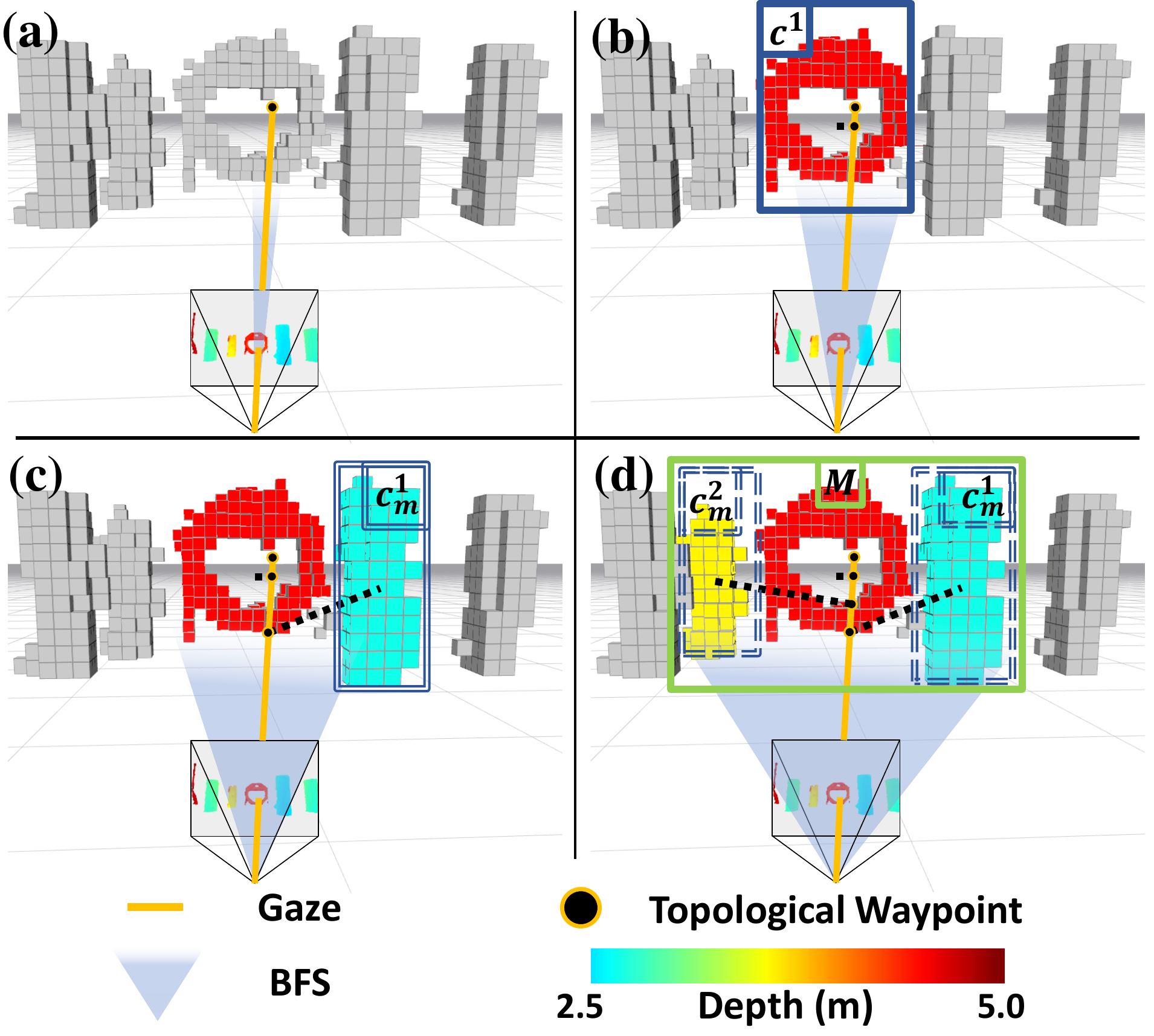}
	\vspace{-0.1cm}
	\captionsetup{font={small}}
	\caption{
		Working principle of our topological path generation algorithm. (a): BFS is conducted on $D$ with gaze as the origin. (b): Object is clustered and signed as $c^{i} (i = 1, 2...)$ if the bounding-box of it contains the gaze point. (c): The first object that does not contain the gaze point is clustered and marked as $c^{1}_{m}$. (d): BFS is continued until the bounding box of $M$ includes the gaze point.
	}
	\label{fig:Topo}
	\vspace{-1.0cm}
\end{figure}

\add{In order to optimize a perception-aware trajectory, we need to firstly generate a topological path that is  consistent with the operator's intention.}
In this section, we propose an intention guiding topological path generation module utilizing gaze fixation. 
An illustration of this section can be found in Fig. \ref{fig:Topo}.
The input of the proposed module is a control signal from RC, a gaze point $g_{I}$ on \add{the first person view (FPV)} image plane \add{$I$} and \delete{the}\add{a} depth image $D$. 
\delete{
	Commonly, the gaze and RC respectively indicate desired direction and speed. 
	Firstly, we have some pre-processes to make the intention more accurate.
	For RC, only \delete{the pitch}\add{one single} channel is utilized and projected to desired speed for trajectory optimization. 
}
For gaze, we register $g_{I}$ to $D$ according to the intrinsic and extrinsic matrix of $D$ and $I$, sign as $g_{D}$. Notably, since $g_{I}$ is a 2-D vector, \delete{so} it cannot be directly projected to 3-D space. Therefore, we use the 
\delete{depth of perception boundary} 
\add{the depth of maximum reliable perception range} as its depth and project it to 3-D space as a local target.
After that, we apply a smoothing filter for the past ten inputs to eliminate the noise of gaze like blink and glance.

\add{After all the pre-processes, we have captured the operator's intention. To ensure that the planned path is consistent with intention, we need to utilize the intention to constrain the topological structure of the path. 
	To achieve this, \textbf{Algorithm}~\ref{alg:TopoPathGen} is proposed. 
	Firstly, $g_{D}$ is utilized as the origin for Breadth First Search (BFS) \cite{bundy1984breadth} on $D$. The BFS would halt once it visits a pixel with its depth in the perception range and is unclustered before.
	Then, we take the pixel as the anchor point for DBSCAN \cite{ester1996density} clustering, resulting in a cluster $c_{temp}$, as illustrated in Fig. \ref{fig:Topo}(b). For each pixel in $c_{temp}$, we label it as clustered. If $g_{D}$ is in the boundingbox of $c_{temp}$, meaning the gaze hit the object, we enqueue $c_{temp}$ into the cluster-set $ \mathcal{C} $ and label it as $c^{i} (i = 1, 2...)$. Otherwise, we check if it is the first object that is not hit by the gaze, if so, we labeled $c_{temp}$ as the first margin cluster $c^{1}_{m}$, as is shown in Fig. \ref{fig:Topo}(c). After assign $c^{1}_{m}$, for each cluster that is not hit by the gaze, we merge it with the margin object as $\mathcal{M}$, like Fig. \ref{fig:Topo}(d) does, and check if $g_{D}$ locates in the bounding-box of $M$. If so, the latest clustered object is signed as $c^{2}_{m}$ and the BFS is stopped.
	Then, we project $g_{D}$ as a line to Body Frame, signed as $g_{B}$. 
	Finally, we project the central points of clusters $ \mathcal{C} = \{ c^{1}, c^{2}...c^{i},  c^{1}_{m}, c^{2}_{m}\}$ to $g_{B}$ as topological waypoints and $g_{B}$ as the topological path.}

\section{Perception-aware Spatial-Temporal Trajectory Optimization}
\label{sec::TrajectoryOPtimization}
\delete{In this section, based on our previous work \cite{wang2021visibility}, we describe an analytical visibility metric that can be easily formulated into a \delete{in}\add{differentiable} cost function.
	Then we propose a trajectory optimizer that refines the path obtained from Sec. \ref{sec::TopoPathGen} into a collision-free and perception-aware trajectory.}

\add{
	In this section, we refine the topological path planned in Sec. \ref{sec::TopoPathGen} to generate a perception-aware trajectory, which is also expected to keep the UAV at the operator's desired speed.
	For perception-awareness, based on our previous work \cite{wang2021visibility}, we describe an analytical visibility metric (Sec. \ref{sec:TrajectoryOPtimization::Visibility}) that can be easily formulated into a differentiable penalty (Sec.~\ref{sec:VisibilityPenalty}).
	For keeping desired speed, we penalize the trajectory execution time (Sec. \ref{sec:ExecutionTimePenalty})
	and limit the speed by constraining the maximum allowed speed (Sec. \ref{sec:DynamicalFeasibilityPenalty}).
	To directly control both the spatial and temporal profile of trajectory, we adopt MINCO \cite{wang2021geometrically} representation (Sec. \ref{sec:TrajectoryOPtimization::Representation}).
}

\subsection{Analytical Visibility Metric}
\label{sec:TrajectoryOPtimization::Visibility}

\begin{figure}[t]
	\vspace{0.0cm}
	\centering
	\includegraphics[width=1\linewidth]{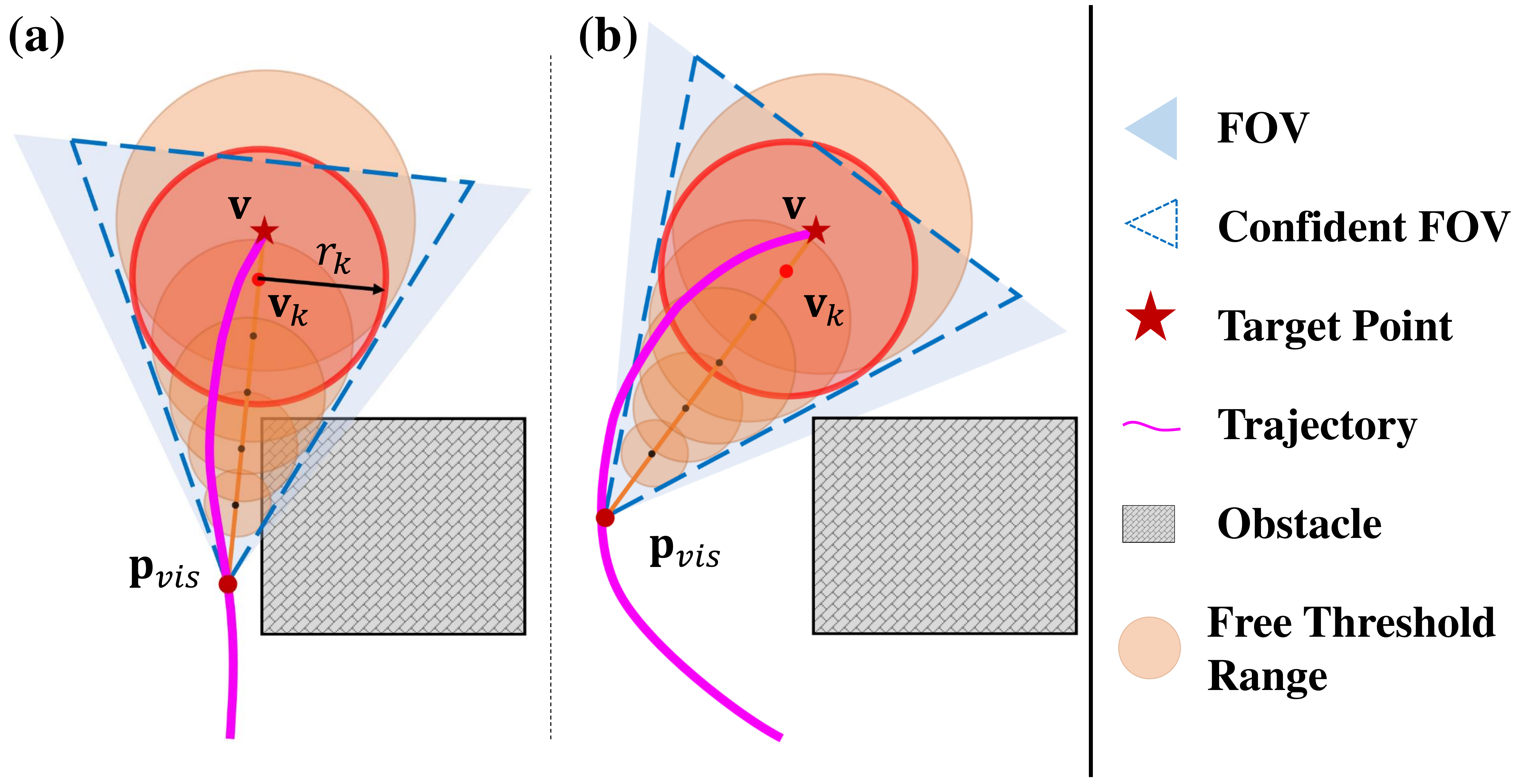}
	\captionsetup{font={small}}
	\caption{
		Illustration of the visibility metric definition. 
		(a): the trajectory ignoring visibility. (b): the trajectory considering visibility.
	}
	\label{fig:visibility}
	\vspace{-0.3cm}
\end{figure}

As the two-dimension profile Fig. \ref{fig:visibility} shows, $\mathbf{p}_{vis}$ is a point on the quadrotor's trajectory and $\mathbf{v}$ is the position of the target which depends on the gaze.
Since the attitude of quadrotor is controlled by the direction of gaze as mentioned in Sec. \ref{sec:Overview}, we assume that the aircraft is always facing the target. 
The blue range represents the FOV of quadrotor.
\delete{
	The blue dashed enveloped area represents the \textbf{confident FOV} where no obstacle is expected.
	However, it is hard and costly to represent this requirement analytically.
}
\add{
	The blue dashed enveloped area represents the \textbf{confident FOV} which is desired to be obstacle-free.
	But it is hard and costly to represent this requirement analytically. 
}
Therefore, we generate a sequence of spherical areas $\{ \mathcal{B}_1,\mathcal{B}_2,...,\mathcal{B}_N \}$ to approximate the confident FOV.  
The $\mathcal{B}_k$ is shown as the red circle, and its center $\mathbf{v}_k$ and radius $r_k$ are calculated by
\begin{equation}
	\label{eq:pi}
	\mathbf{v}_k = \mathbf{p}_{vis} + \psi_k(\mathbf{v}-\mathbf{p}_{vis}),~~
	r_k = \rho \cdot \psi_k \cdot ||{\mathbf{v}-\mathbf{p}_{vis}}||,
\end{equation}
where $\psi_k = k/N, k\in\{1,2,...,N\}$, and $\rho$ is a constant parameter that determines size of the confident FOV. 

\delete{
	Then we can enhance the visibility of target analytically for each ball by $\Xi(\mathbf{v}_k) > r_k$, 
	where $\Xi(\mathbf c_i): \mathbb R^3 \rightarrow \mathbb{R}$ is the distance to the closest obstacle.
	We transform the requirement into}

\delete{
	Based on previous work \cite{wang2021visibility}, we improve \add{the} penalty function of visibility with Eq. \ref{eq:visbility}, which will be detailed in  \ref{sec:TrajectoryOPtimization::Formulation}.
	As shown in Fig. \ref{fig:visibility}(b), by squeezing out obstacles inside the spherical areas, we effectively increase the visibility of the target. 
}

\add{
	Then we can enhance the visibility of target analytically 
	by requiring each ball to satisfy
	\begin{equation}
		\label{eq:visbility_iros}
		\Xi(\mathbf{v}_k) > r_k,
		~~\Xi(\mathbf c_i): \mathbb R^3 \rightarrow \mathbb{R},
	\end{equation}
	which is used to construct visibility penalty function in \cite{wang2021visibility} and $\Xi(\mathbf c_i)$ is the distance to the closest obstacle.
	However, in this paper we transform the requirement of visibility Eq.\ref{eq:visbility_iros} into
	\begin{equation}
		\label{eq:visbility}
		(\Xi(\mathbf{v}_k)/l_k) > \rho, ~~
		l_k = \psi_k ||{\mathbf{v}-\mathbf{p}_{vis}}||,
	\end{equation}
	based on which we design the visibility penalty in this work.
	This penalty performs better than \cite{wang2021visibility} in terms of enhancing visibility, which is detailed in Sec. \ref{sec:VisibilityPenalty}.
	
	As shown in Fig. \ref{fig:visibility}(b), by expelling obstacles from the spherical area, this method effectively increase the visibility of the target. 
	
}

\subsection{Trajectory Representation}
\label{sec:TrajectoryOPtimization::Representation}
In this paper, we adopt the MINCO \cite{wang2021geometrically} to conduct spatial-temporal deformation of the flat-output trajectory.
An $s$-order MINCO trajectory is indeed a 2$s$-order polynomial
spline with constant boundary conditions, which is defined as:
\begin{align}
	\begin{split}
		\mathfrak{T}_{\mathrm{MINCO}} = \Big\{&p(t):[0, T]\mapsto\mathbb{R}^m \Big|~\mathbf{c}=\mathcal{M}(\mathbf{q},\mathbf{T}),~\\
		&~~\mathbf{q}\in\mathbb{R}^{m(M-1)},~\mathbf{T}\in\mathbb{R}_{>0}^M\Big\},
	\end{split}
\end{align}
where $p(t)$ is an $m$-dimensional $M$-piece polynomial trajectory.
$\mathbf{c} = (\mathbf{c}_1\tp, \dots, \mathbf{c}_M\tp)\tp \in \mathbb{R}^{2Ms \times m}$ is the polynomial coefficient and $\mathcal{M}(\mathbf{q},\mathbf{T})$ is a linear-complexity smooth map from intermediate points $\mathbf{q}$ and a time allocation $\mathbf{T}$ for all pieces to the coefficients of splines.
A spline with $\mathbf{c}=\mathcal{M}(\mathbf{q},\mathbf{T})$ is exactly the unique control effort minimizer of an s-integrator that passes $\mathbf{q}$.
For any penalty function $F(\mathbf{c},\mathbf{T})$ with available gradients, MINCO can also serve as a linear-complexity differentiable layer $H(\mathbf{q},\mathbf{T})=F(\mathcal{M}(\mathbf{q},\mathbf{T}), \mathbf{T})$.
Then $\partial H/\partial\mathbf{q}$ and $\partial H/\partial\mathbf{T}$ can be obtained efficiently from the corresponding $\partial F/\partial\mathbf{c}$ and $\partial F/\partial\mathbf{T}$.

The $i$-th piece $p_i(t)$ of trajectory is defined by
\begin{equation}
	\label{equ:polytraj}
	p(t)=p_i(t-t_{i-1}),~\forall t\in[t_{i-1},t_i].
	\vspace{-0.4cm}
\end{equation}

\begin{equation}
	p_i(t)=\mathbf{c}_i\tp\beta(t),~\forall t\in[0,T_i],
\end{equation}
where $\beta(t):=[1, t, \cdots, t^N]\tp$ is the natural basis, $\mathbf{c}_i$ the coefficient matrix, $T_i=t_i-t_{i-1}$ and $T = \sum_{i=1}^{M} T_i$. 

Specifically, we define \textbf{constraint points} $\mathring{\mathbf{p}}_{i,j}$ which are sampled on each piece of the trajectory by
$\mathring{\mathbf{p}}_{i,j} = p_i(({j}/{\kappa_i})T_i),~
j \in \{0, 1,2,..., \kappa_i\}$,
where $\kappa_i$ is the sample number on the $i$-th piece. 
\add{These points are used to transform the time integral of some constraint functions, such as the Dynamical Feasibility Penalty in Sec. \ref{sec:DynamicalFeasibilityPenalty}, into weighted sum of sampled penalty functions.}
\delete{The constraint points are used to transform the time integral of some constraint function, such as dynamical feasibility constraint, into weighted sum of sampled penalty function.}

\subsection{Problem Formulation}
\label{sec:TrajectoryOPtimization::Formulation}
Inspired by \cite{zhou2020raptor}, first we generate an intermediate warm-up trajectory as the initial trajectory utilizing the topological path from Sec. \ref{sec::TopoPathGen}.
Then taking our requirements into account, we formulate the trajectory optimization problem as
\begin{equation}
	\label{eq:objfun}
	\min_{\mathbf{q},\mathbf{T}}~ {[J_e, J_d, J_t, J_{vis}, J_c, J_u] \cdot \mathbf{\lambda}},
\end{equation}
where $\mathbf{\lambda}$ is the a weighting vector to trade off each penalty which will be described below. In this paper, we use the quasi-Newton method proposed in \cite{lewis2013nonsmooth} which supports nonsmooth cost functions to solve this unconstrained optimization problem.


\subsubsection{Execution Time Penalty $J_t$}
\label{sec:ExecutionTimePenalty}
\delete{
	Since quadrotor is expected to fly at maximum velocity to the full extent.
}
\add{
	To increase the speed of UAV, we penalty the total execution time 
	$J_t=\sum_{i=1}^{M} T_i$.
}
Its gradient is
${\partial J_t}/{\partial \mathbf{c}} = \mathbf{0},
{\partial J_t}/{\partial \mathbf{T}} = \mathbf{1}$.

\subsubsection{Dynamical Feasibility Penalty $J_d$}
\label{sec:DynamicalFeasibilityPenalty}
\delete{
	In order to make the operator feel that they are effectively controlling the quadrotor, the maximum allowed velocity of quadrotor is determined by the remote control input. 
}
\add{
	To prevent the speed from increasing excessively introduced by $J_t$, we use the desired speed to set the value of the maximum allowed speed.
}
Additionally violent shifts of speed will lead to a blurred first view, which is unfriendly to operator.
Thus not only velocity, we also limit the amplitude of acceleration and jerk.
The constraints are written by
\begin{equation}
	\label{eq:Jd_cost}
	\mathcal{G}_* = {p_i^{(n)}}(t)^2 - max_*^2, ~~
	\forall{t}  \in [0, T_i],
\end{equation}
where $*=\{v,a,j\}$, $n=\{1,2,3\}$, and $t=jT_i/\kappa_i$, respectively. 
$max_v, max_a, max_j$ are maximum allowed velocity, acceleration and jerk.
Inspired by the method in \cite{jennings1990computational}, the dynamical feasibility penalty is obtained by computing weighted sum of sampled constraint function:
\begin{equation}
	\label{eq:Jd_sum}
	(J_d)_*=
	\sum_{i=0}^{M}
	\frac{T_i}{\kappa_i}\sum_{j=0}^{\kappa_i}\bar{\omega}_j\max{(\mathcal{G_*}(\mathring{\mathbf{p}}_{i,j}),\mathbf{0})^3},
\end{equation}
where  $(\bar{\omega}_0,\bar{\omega}_1,\dots,\bar{\omega}_{\kappa_i-1},\bar{\omega}_{\kappa_i})=(1/2,1,\cdots,1,1/2)$ are the quadrature coefficients following the trapezoidal rule \cite{press2007numerical} and $\mathring{\mathbf{p}}_{i,j}$ are constrain points. 
The gradient can be written by
\begin{equation}
	\label{eq:jd_c}
	\frac{\partial (J_d)_*}{\partial \mathbf{c}_i} = \frac{\partial (J_d)_*}{\partial \mathcal{G}} \frac{\partial \mathcal{G}}{\partial \mathbf{c}_i},
\end{equation}
\begin{equation}
	\label{eq:jd_t}
	\frac{\partial (J_d)_*}{\partial T_i} = 
	\frac{\partial (J_d)_*}{\partial \mathcal{G}} \frac{\partial \mathcal{G}}{\partial t} \frac{\partial t}{\partial T_i}+
	\sum_{j=0}^{\kappa_i}\frac{\bar{\omega}_j}{\kappa_i}{\kappa_i}\max{(\mathcal{G_*}(\mathring{\mathbf{p}}_{i,j}),\mathbf{0})^3},
\end{equation}
\begin{equation}
	\label{eq:Jd_gradient}
	\frac{\partial \mathcal{G}_*}{\partial \mathbf{c}_i}=2{\beta^{(n)}(t)} {p_i^{(n)}}(t)\tp, ~~
	\frac{\partial \mathcal{G}_*}{\partial t}=2{\beta^{(n+1)}}(t)\tp \mathbf{c}_i p_i^{(n)}(t),
\end{equation}

\subsubsection{Visibility Penalty $J_{vis}$}
\label{sec:VisibilityPenalty}
To achieve the visibility demand defined in Sec. \ref{sec:TrajectoryOPtimization::Visibility}, we design visibility penalty as
\begin{equation}
	\label{eq:vis_penalty}
	J_{vis}=\sum_{i=0}^{M}\frac{T_i}{\kappa_i}\sum_{j=0}^{\kappa_i}\bar{\omega}_j
	F_{vis}(\mathring{\mathbf{p}}_{i,j}).
\end{equation}
where these constraint points $\mathring{\mathbf{p}}_{i,j}$ are selected as the visibility points $\mathbf{p}_{vis}$. For each visibility point $\mathbf{p}_{vis}$, we define
\begin{equation}
	\label{eq:vis_func}
	F_{vis}(\mathbf{p}_{vis})=\sum_{k=1}^{N}\max{(f_{vis}((\mathbf{v}_k), 0)}^3,
\end{equation}
\begin{equation}
	\label{eq:vis_func_func}
	f_{vis}(\mathbf{v}_k)=(\rho - \Xi(\mathbf{v}_k)/l_k),
\end{equation}
where $N, \mathbf{v}_k, l_k$ and $\rho$ are the parameters of spherical areas mentioned in Sec. \ref{sec:TrajectoryOPtimization::Visibility} and $\Xi(\mathbf{v}_k)$ is obtained from ESDF. 
\delete{
	Compared to our previous work \cite{wang2021visibility}, this penalty function reduces the direct impact of $l_k$ changes and focuses more on decreasing occlusion.
}

Then the gradient of $f_{vis}$ cost can be calculated by
\begin{align}
	\label{eq:vis_func_grad}
	\frac{\partial{f_{vis}}}{\partial{\mathbf{p}_{vis}}}= 
	\frac{1}{l_k^2}
	\left[\frac{\psi_k}{l_k}\Xi(\mathbf{v}_k)(\mathbf{p}_{vis}-
	\mathbf{v}_k)-
	(1-\psi_k) l_k \frac{\partial{\Xi{(\mathbf{v}_k)}}}{\partial{\mathbf{v}_k}}
	\right],
\end{align}
where $\partial{\Xi{(\mathbf{v}_k)}}/{\partial{\mathbf{v}_k}}$ can be efficiently acquired from ESDF.

Note that each visibility point  $\mathbf{p}_{vis}$ is a constraint point $\mathring{\mathbf{p}}_{i,j}=p_i(t)$, $t=jT_i/\kappa_i$, whose gradient is written by
\begin{equation}
	\label{eq:p_gand}
	\frac{\partial \mathring{\mathbf{p}}_{i,j}}{\partial \mathbf{c}_i}
	={\beta(t)}, ~~
	\frac{\partial \mathring{\mathbf{p}}_{i,j}}{\partial t}
	={ \dot \beta}(t)\tp \mathbf{c}_i.
\end{equation}
Then with the above Eq. \ref{eq:vis_func_grad}, \ref{eq:p_gand} the gradient of the visibility penalty $J_{vis}$ can be calculated easily using the chain rule.

\add{
	\begin{figure}[t]
		\centering
		\includegraphics[width=0.9\linewidth]{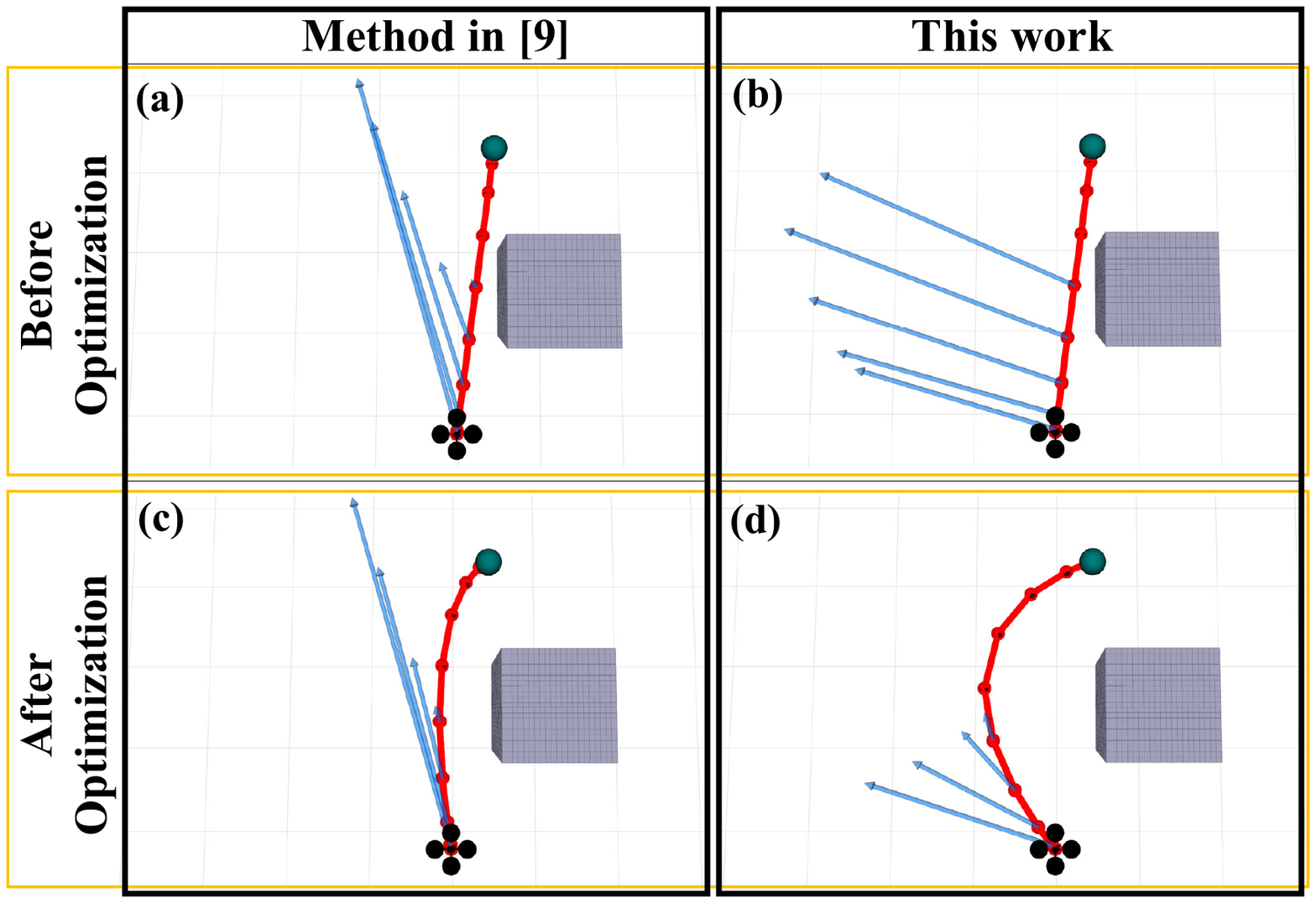}
		\vspace{-0.2cm}
		\captionsetup{font={small}}
		\caption{
			\add{
				Comparison of \cite{wang2021visibility} and this work. 
				The gray square is an obstacle, the black quadrotor is the initial position of the drone, the green ball indicates the target, the red curve indicates the trajectory and the points on it indicate the constraint points, and the light blue arrows indicate the negative gradients of the visibility costs.
			}
		}
		\label{fig:metric_compare}
		\vspace{-1.0cm}
	\end{figure}
	
	Notably, in this work, we upgrade the formulation of the visibility requirements compared to \cite{wang2021visibility}.
	Specifically, compared to Eq. \ref{eq:visbility}, the method in \cite{wang2021visibility} formulates the visibility requirement into $\Xi(\mathbf{v}_k) > r_k$, in which the visibility cost for each visibility  point is designed as
	\begin{equation}
		\vspace{-0.2cm}
		\label{eq:iros_vis_func}
		\widetilde{F}_{vis}(\mathbf{p}_{vis})=\sum_{k=1}^{N}\max{(\widetilde{f}_{vis}((\mathbf{v}_k), 0)}^3,
		\vspace{-0.1cm}
	\end{equation}
	\begin{equation}
		\label{eq:iros_vis_func_func}
		\widetilde{f}_{vis}(\mathbf{v}_k)=
		({r_k}^2 - {\Xi(\mathbf{v}_k)}^2).
		\vspace{-0.1cm}
	\end{equation}
	Comparing Eq. \ref{eq:vis_func_func} and Eq. \ref{eq:iros_vis_func_func}, the key improvement is that in Eq. \ref{eq:vis_func_func} the direct effect of $||{\mathbf{v}-\mathbf{p}_{vis}}||$ on the cost function is weakened, which makes the cost more focused on occlusion, rather than directly shortening $||{\mathbf{v}-\mathbf{p}_{vis}}||$ to reduce the cost. 
	
	To show the significance of the improvement intuitively, in Fig. \ref{fig:metric_compare} we visualize these 
	two visibility cost functions' negative gradients, which are $-{\partial{\widetilde{F}_{vis}}}/{\partial{\mathbf{p}_{vis}}}$ and $-{\partial{F_{vis}}}/{\partial{\mathbf{p}_{vis}}}$.
	We use these two costs separately for comparison in the trajectory optimization problem in Sec.~\ref{sec:TrajectoryOPtimization::Formulation}.
	Both of them are set up with the same optimization parameters, and each penalty functions of the optimization is the same except for the visibility penalty.
	To fairly compare these two methods, two visibility costs are adjusted to the same order of magnitude.

	As shown in Fig. \ref{fig:metric_compare}(a) and Fig. \ref{fig:metric_compare}(c), the gradient of visibility cost in \cite{wang2021visibility} has a very large component in the direction of $({\mathbf{v}-\mathbf{p}_{vis}})$, which leads to the trajectory after optimization with less effective visibility improvement.
	As shown Fig. \ref{fig:metric_compare}(b) and Fig. \ref{fig:metric_compare}(d), compared to \cite{wang2021visibility}, the visibility cost in this work can enhance visibility more remarkably.}

\delete{
	\subsubsection{Collision Avoidance Penalty $J_c$}
	We adopt the collision evaluation of our previous work EGO-Planner to guide the trajectory with a collision-free path. 
	Readers can refer to \cite{zhou2020egoplanner} and \cite{zhou2021decentralized} for more detailed explanations.
}

\delete{
	\subsubsection{Uniform Distribution Penalty $J_u$}
	Since our obstacle avoidance is constrained on discrete constraint points, non-uniform distribution may cause the trajectory to pass through thin obstacles.
	The constraint points are expected to \delete{equally space-uniform} uniformly spaced.  
	More details can be found in \cite{zhou2021decentralized}.
}

\subsubsection{Other Penaltise}
\add{
	In order to guide trajectory with a collision-free path, we adopt the collision evaluation of EGO-Planner \cite{zhou2020egoplanner} to construct Collision Avoidance Penalty $J_c$. 
	We also formulate Control Effort $J_e$ and Uniform Distribution Penalty $J_u$ to make trajectory smooth and reliable.
	Readers can refer to \cite{zhou2020egoplanner} and \cite{zhou2021decentralized} for more detailed explanations.
}

\section{Evaluation and Experiment}
\label{sec:Results}

\subsection{Implementation Details}
\label{sec:ImpleDetail}

All real-world evaluations are completed by multiple volunteers and most of them have little flight experience.
The experimental procedure is as follows.
Upon arrival at the experiment site, participants are informed of the task. Then they are invited to practice three times before each experiment.
Finally, they are asked to perform each experiment three times repeatedly and the relevant data is recorded.

Our quadrotor system's hardware setups is based on our previous work \cite{zhou2020egoplanner}.
The parameters of visibility metric are $N=20,~\rho=0.8$.
\add{Our method can run in real	time and cost about 5 ms to solve the optimization problem formulated in Sec. \ref{sec::TrajectoryOPtimization} on an onboard computer DJI Manifold 2C. Readers can get a better understanding of each experiment from the attached video\footnote{https://www.youtube.com/watch?v=WYujLePQwB8}.
}

\add{
	\subsection{Evaluation for Keeping Desired Speed}
	\label{sec:EvaSpeed}	
	\begin{figure}[t]
		\centering
		\includegraphics[width=1\linewidth]{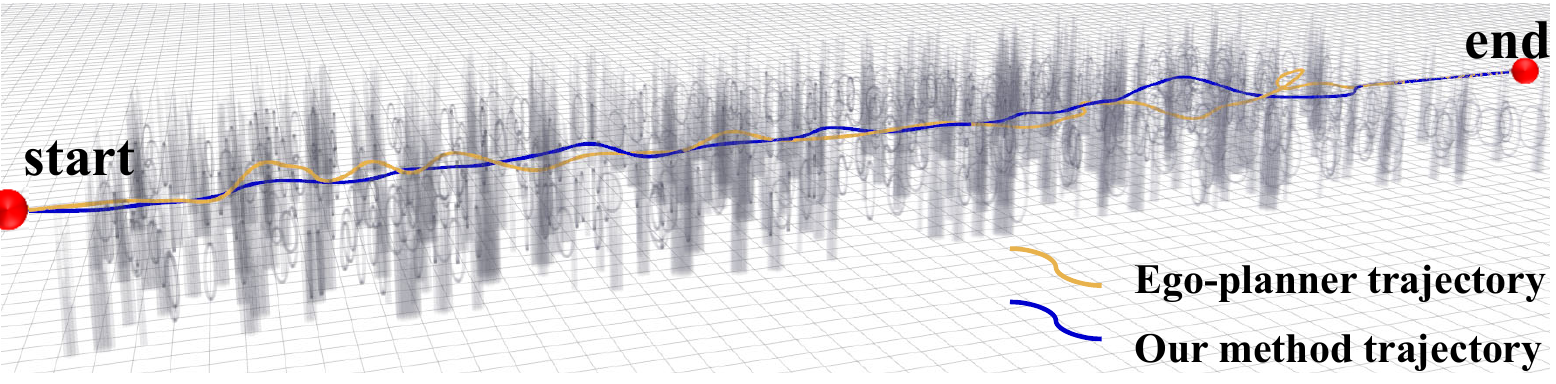}
		\vspace{-0.6cm}
		\captionsetup{font={small}}
		\caption{
			\add{
				Comparison of the trajectory of Ego-planner and our method in a complex environment, which has an obstacle density of 0.625 obs/$m^2$.
				The size of the environment is $60 \times 20 \times 2m$.
			}
		}
		\vspace{-0.3cm}
		\label{fig:minco_env}
	\end{figure}
	
	\begin{figure}[t]
		\centering
		\includegraphics[width=1\linewidth]{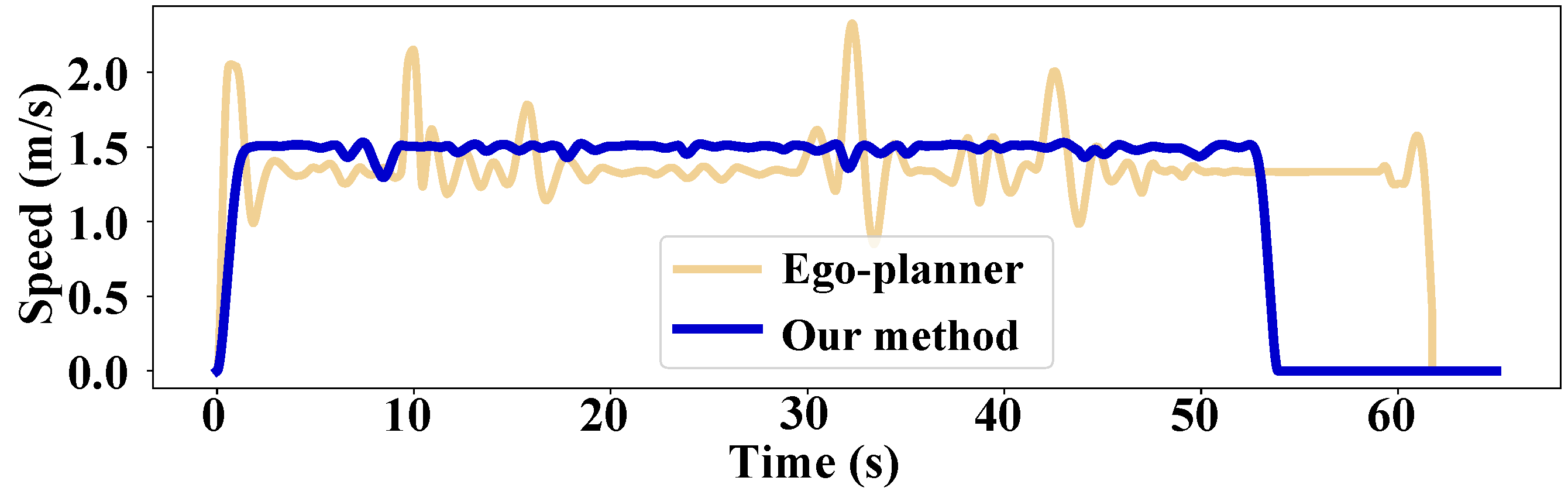}
		\vspace{-0.6cm}
		\captionsetup{font={small}}
		\caption{
			\add{
				Speed comparison between Ego-planner and our method.
			}
		}
		\vspace{-1.4cm}
		\label{fig:minco_result}
	\end{figure}
	
	To show the significance of the improvement in our trajectory optimization for maintaining the operator's desired speed, as Fig \ref{fig:minco_env} shows, we compare our method with Ego-planner \cite{zhou2020egoplanner}, which is the basis of the trajectory planning in \cite{wang2021visibility}.
	We set the same start and end points, and the same desired speed as 1.5 m/s.
	Since only the performance of keeping expected speed is compared, in this experiment, we set the Visibility Penalty (Sec. \ref{sec:VisibilityPenalty}) to 0 in our method.
	
	As Fig. \ref{fig:minco_result} shows, with the aids the newly added Execution Time Penalty (Sec. \ref{sec:ExecutionTimePenalty}) and Dynamical Feasibility Penalty (Sec. \ref{sec:DynamicalFeasibilityPenalty}), our method performs remarkably better at keeping desired speed.
	However, Ego-planner often requires a change in speed for obstacle avoidance due to the limitations of the uniform B-spline curve.
	
	\delete{
		Frequent uncontrolled speed changes will make the operator feel like they are losing effective control of the drone, which is not friendly to operator.
	}
}

\subsection{Evaluation for Gaze Enhanced Intention Capture}
\label{sec:EvaTopo}

\begin{figure}
	\centering
	\includegraphics[width=0.9\linewidth]{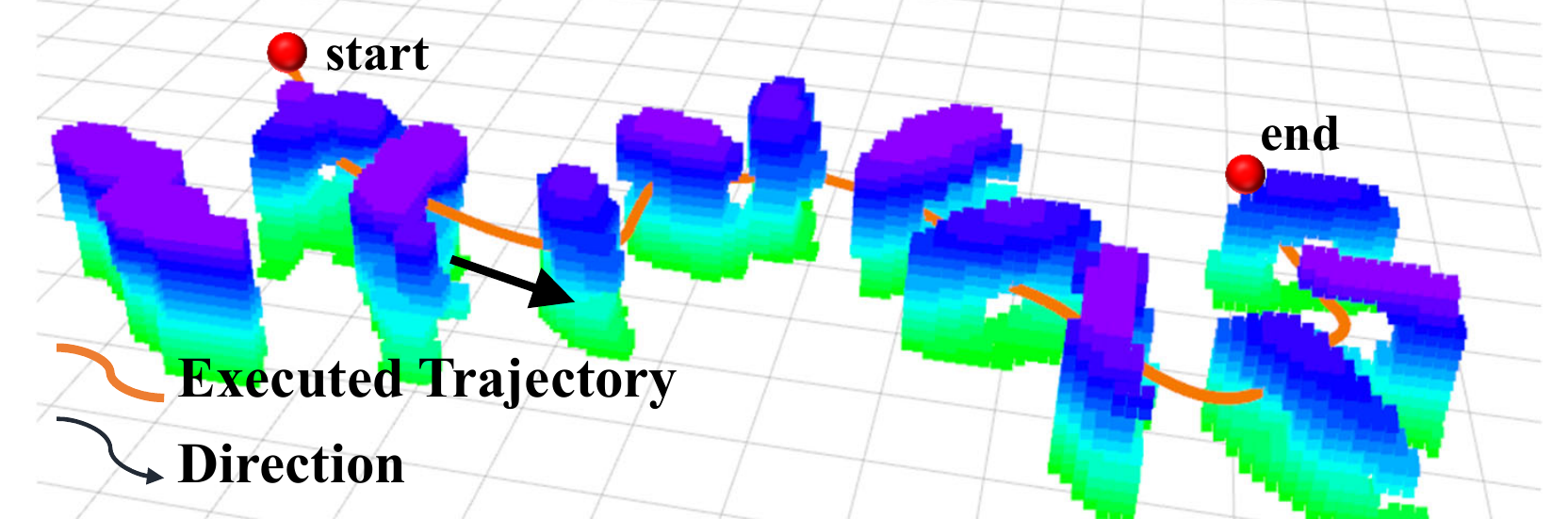}
	\vspace{-0.1cm}
	\captionsetup{font={small}}
	\caption{
		Trajectories executed by the quadrotor and online generated map which is colored by height in drone racing experiment.
	}
	\label{fig:topo_rviz}
	\vspace{-0.4cm}
\end{figure}

To demonstrate the significance of gaze in intention inference and topological path generation, we implement this experiment in a challenging drone racing scenario as shown in Fig. \ref{fig:cover} and Fig. \ref{fig:topo_rviz}. We compare our gaze enhanced method with the commonly used RC method: assistive teleoperation system which captures intention directly from RC. 
\add{
	Note that both methods use the same perception-aware trajectory optimization method which is described in Sec. \ref{sec::TrajectoryOPtimization}.
}

To evaluate the stability and accuracy of intention inference, we proposed three independent indicators: average success rate of passing rings $\overline{S}$, topological stability $\mathcal{T}_{s}$ and average finish time $\overline{T}$. 
We define $\overline{S}$ and $\mathcal{T}_{s}$ as
\begin{equation}
	\begin{aligned}
		\overline{S} = \frac{N_{success}}{N_{sum}},~~
		\mathcal{T}_{s} = \frac{N_{sum}}{\sum_{i=0}^{N_{sum}} (N_{switch})_{i}},
		\label{con:succrate} 
	\end{aligned}
	\vspace{-0.2cm}
\end{equation}
where \delete{$(N_{switch})_{i}$ means the times that the topological structure of the planned path inconsistent with intention when passing through the $i$-th ring.} \add{$(N_{switch})_{i}$ counts times the topology of the planned path is inconsistent with the intent as it passes through the $i$-th ring.} $N_{success}$ is the number of rings that quadrotor passes through successfully and $N_{sum}$ is the total number of rings.

\begin{figure}
	\centering
	\includegraphics[width=0.9\linewidth]{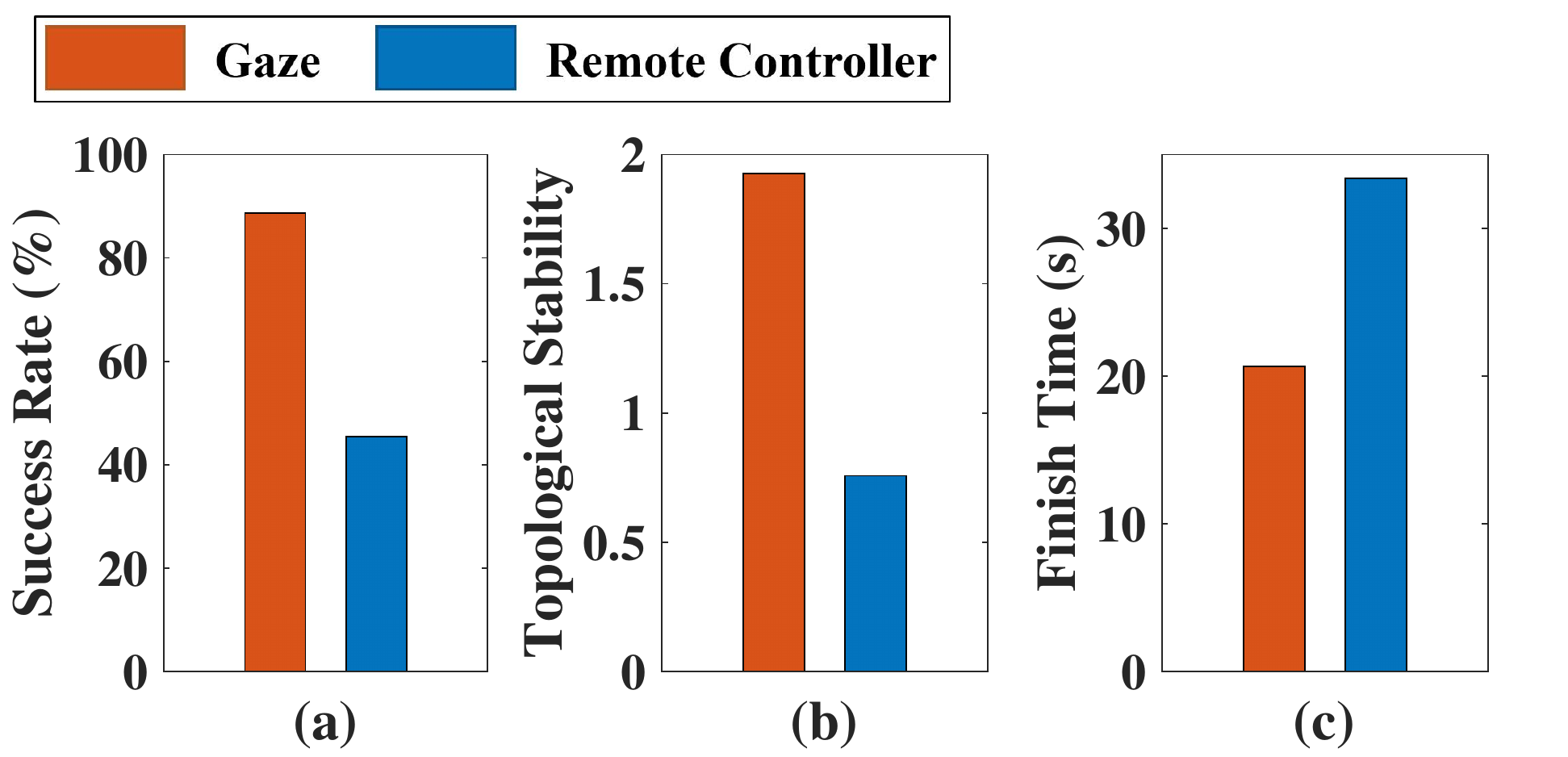}
	\captionsetup{font={small}}
	\vspace{-0.1cm}
	\caption{
		Results of the comparisons in drone racing experiment.
	}
	\label{fig:topo_hist}
	\vspace{-1.8cm}
\end{figure}

As the results show in Fig. \ref{fig:topo_hist}, benefit from gaze, our method capture precise intention, which helps us have better topological stability and a higher success rate.
However, for the RC method, inaccurate intention causes operators to adjust the RC input repeatedly.
This leads to low topological stability, which is the major cause of the low success rate.


\begin{figure*}[ht]
	\centering
	\includegraphics[width=0.95\textwidth]{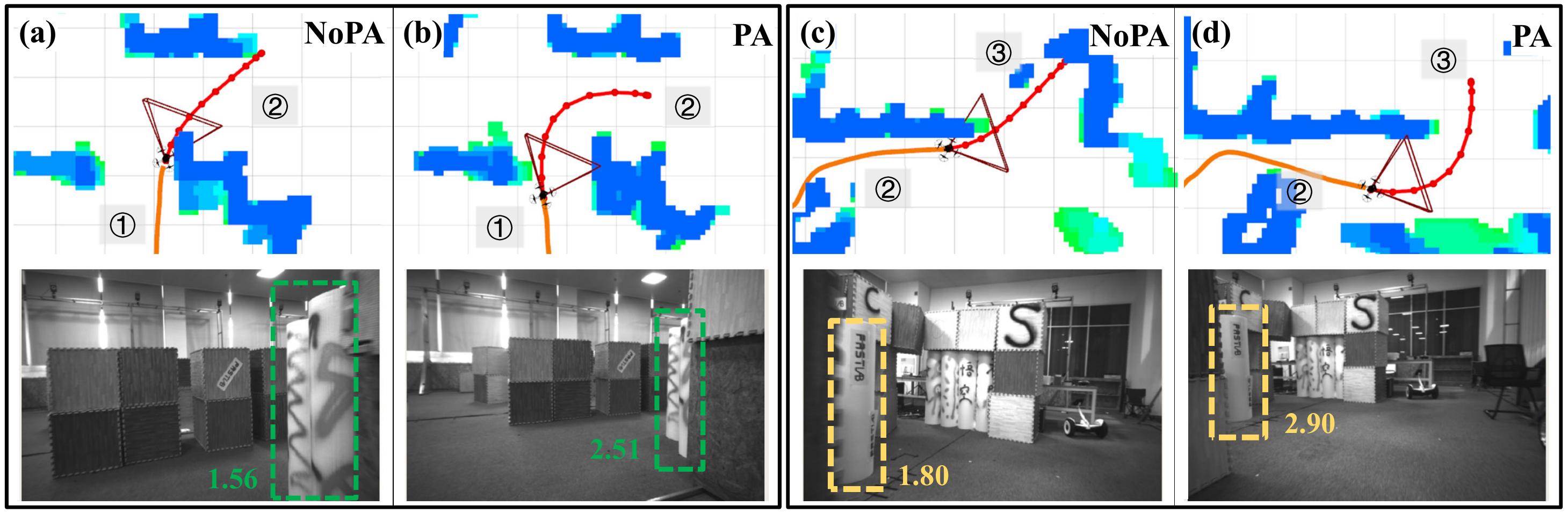}
	\vspace{-0.1cm}
	\caption{
		Comparison of PA (perception-aware) and NoPA (without perception-aware). 
		\textbf{Top}: the visualization of the experiment. 
		The orange curve denotes the executed trajectory, the red curve denotes the planned trajectory, and the red pyramid stands for FOV.
		(a) and (b) show the quadrotor flies from \ding{172} to \ding{173}.
		(c) and (d) show the quadrotor flies from \ding{173} to \ding{174}.
		\textbf{Bottom}: the first-person perspective of the quadrotor, the obstacles marked by green and yellow are corresponding to to SO$_1$ (surprise obstacle) and SO$_2$ in Fig. \ref{fig:pa_tpv}.
		The colored numbers indicate the distance between the quadrotor and the obstacles.
	}
	\label{fig:pa_rviz}
	\vspace{-0.6cm}
\end{figure*}

\subsection{Evaluation for Perception-aware Trajcetory Optimization}
\label{sec:EvaPA}

To validate the superiority of the proposed perception-aware trajectory optimization method for flight safety, we compare it with the one \delete{ does not equip with} \add{which is not equipped with} this feature. 
And \textbf{PA} and \textbf{NoPA} are used to refer to whether the perception-aware method is used in our system \add{or not}.

\begin{figure}
	\centering
	\includegraphics[width=0.95\linewidth]{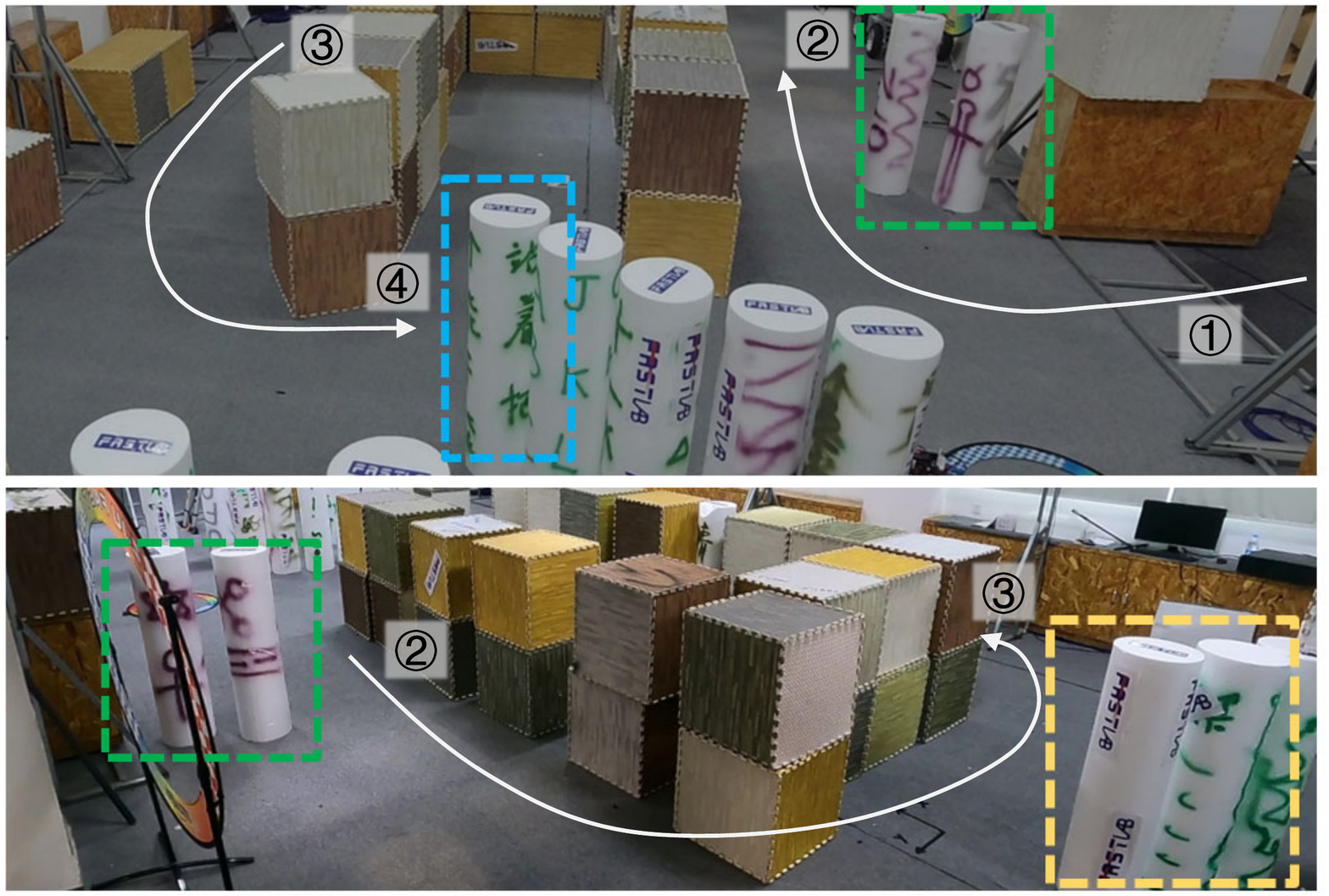}
	\captionsetup{font={small}}
	\caption{
		Two third-person views of the experiment.
		Volunteers are asked to fly the quadrotor from \ding{172} to \ding{175}, and through \ding{173} and \ding{174} in turn.
		SO$_1$,  SO$_2$ and  SO$_3$ are  marked in green, yellow and blue respectively.
	}
	\label{fig:pa_tpv}
	\vspace{-1.2cm}
\end{figure}

The volunteers are asked to fly quadrotor following the white route (from\circled{1}to\circled{4}) in the maze, as Fig. \ref{fig:pa_tpv} shows.
There are three corners where we place \textbf{surprise obstacles} (\textbf{SO}) right behind, and these surprise obstacles (SO$_1$, SO$_2$ and SO$_3$) are marked by green, yellow and blue boxes respectively \add{in Fig. \ref{fig:pa_tpv} and Fig. \ref{fig:pa_rviz}}.
As Fig. \ref{fig:pa_rviz} shows, PA always turns on a safer trajectory that enhances the visibility of the target, which allows the quadrotor to see SO earlier.

\begin{figure}
	\centering
	\includegraphics[width=0.9\linewidth]{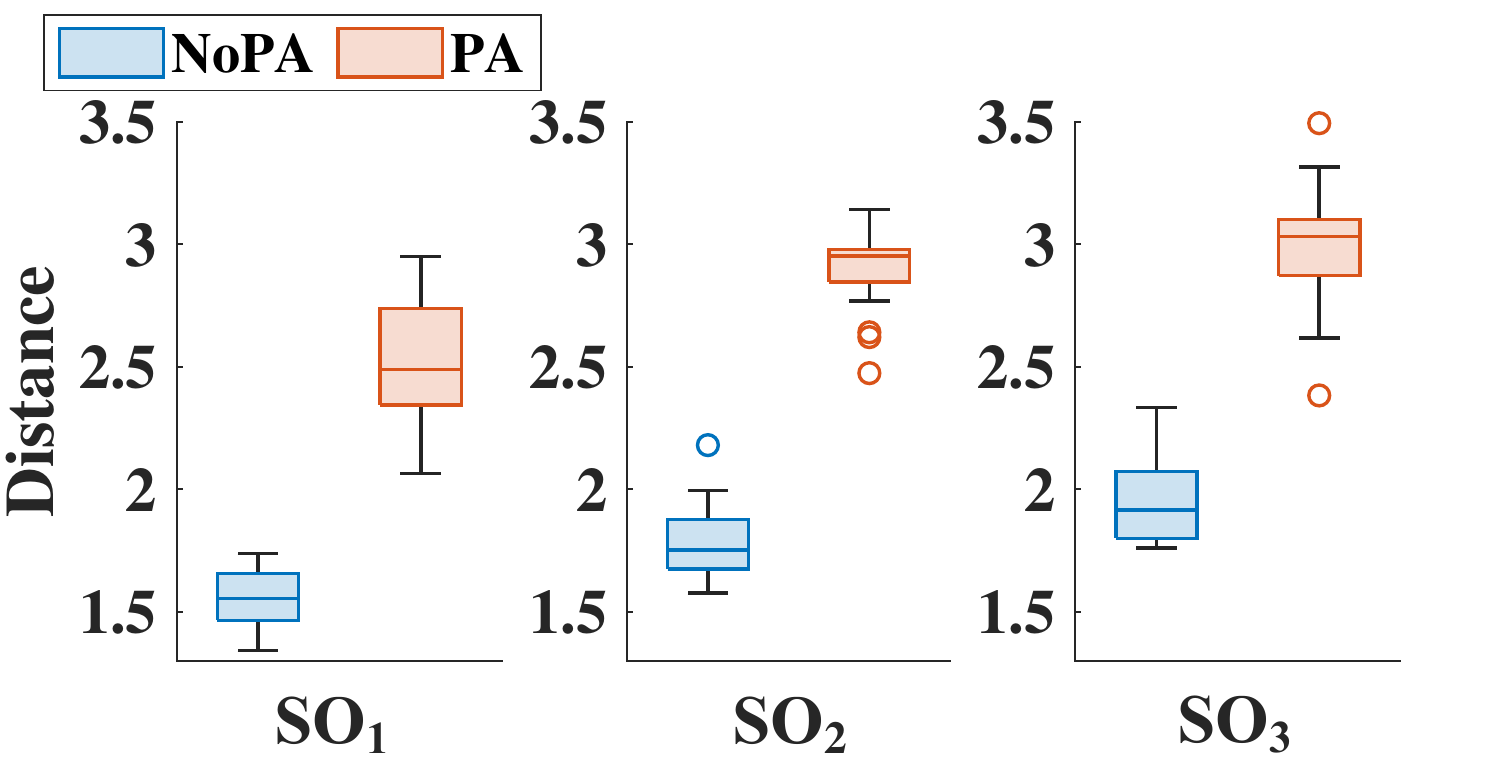}
	\captionsetup{font={small}}
	\vspace{-0.2cm}
	\caption{
		Results of comparisons between NoPA and PA, the indicator refers to the distance that SO first appears fully in the field of view.
	}
	\label{fig:pa_box}
	\vspace{-2.2cm}
\end{figure}

Additionally, we record the distance between SO and the quadrotor when volunteers see SO for the first time.
In Fig. \ref{fig:pa_box}, the results presented as box charts show that our PA method significantly increases the distance between SO and the quadrotor when SO is seen the first time, which demonstrates that our method assists operators to fly safer and makes more time for operators to do decisions.

\delete{The results are presented as box charts in Fig. \ref{fig:pa_box}. The results show that our perception-aware method significantly increases the distance between SO and the quadrotor when SO is seen the first time, which demonstrates that our method assists operators to fly more safely and make more timely decisions.}
\delete{We refer the readers to the attached video for a more intuitive awareness of this experiment.}

\subsection{System-level Comparison}
\label{sec:SysExp}
In this section, experimental scenarios have been divided into indoor flight and outdoor flight. 
Although there are some previous works on assistive teleoperation \cite{yang2020assisted,yangintention,israelsen2014automatic}, none of them have runnable software available.
To prove that our system is practical and robust, we compare our system with the industry-leading consumer UAV, DJI Mavic Air 2\footnote{https://www.dji.com/cn/mavic-air-2}, which is equipped with three-direction perception, and DJI's APAS 3.0 (Advanced Pilot Assistance Systems).
Note that Mavic Air 2 obtain operator's intention from the input of RC only.

\subsubsection{Indoor Flight Test}
\label{sec:ExpIndoor}
\begin{figure}
	\centering
	\includegraphics[width=\linewidth]{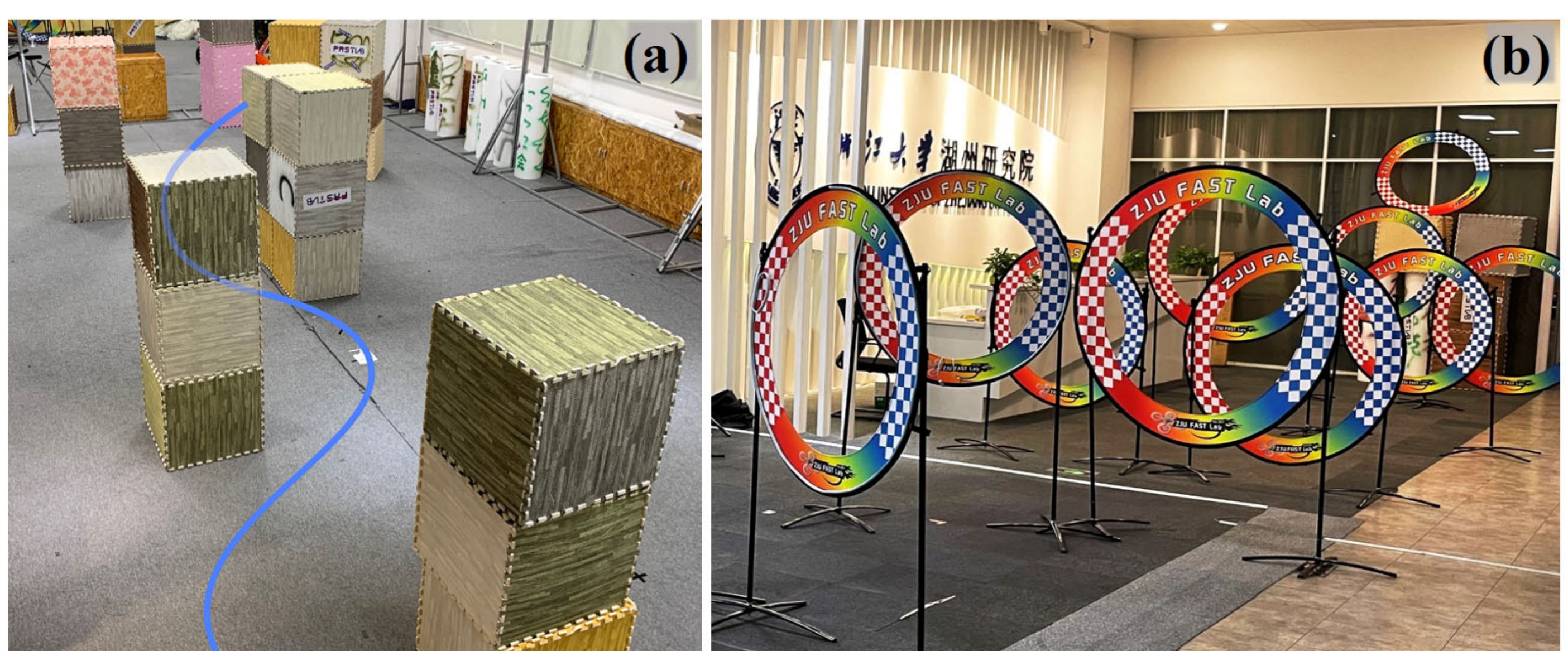}
	\vspace{-0.3cm}
	\captionsetup{font={small}}
	\caption{
		Two third-person views of the indoor experiment.
		(a): the first half of the experiment. The blue line is the prescribed topological path.
		(b): the second half of the experiment.
	}
	\vspace{-0.4cm}
	\label{fig:indoor_tpv}
\end{figure}

We conduct challenging indoor flight experiments, as shown in Fig. \ref{fig:indoor_tpv}, to prove the high controllability and robustness of our system. 
\delete{Readers can get a better understanding of the experiment scene from the attached video.}

In the first half of the experiment, as Fig. \ref{fig:indoor_tpv}(a) shows, the obstacle is densely placed with less than 1.5m away from each other. The task of the scene is to achieve the goal, $11m$ away from the start point, within the prescribed topological path as soon as possible. The results are presented in TABLE~\ref{tab:IndoorSysComp}. Both the average time consumption and maximum speed of our system outperform Mavic Air 2 significantly.
For Mavic Air 2, redundant sensors and limited computing power lead to a conservative performance.
Furthermore, this RC-only assistive teleoperation system cannot obtain precise intention, which is unfriendly to operator in a dense environment. 
\delete{The consumer UAV often wrongly infers the topological intention hard to control in a dense environment even with low speed and APAS, which can be found in the attached video. 
	In comparison, it is pretty easy for all volunteers to control our gaze-enhanced system with a velocity up to 1.5 m/s after practicing less than three times.}

\begin{figure}
	\centering
	\includegraphics[width=\linewidth]{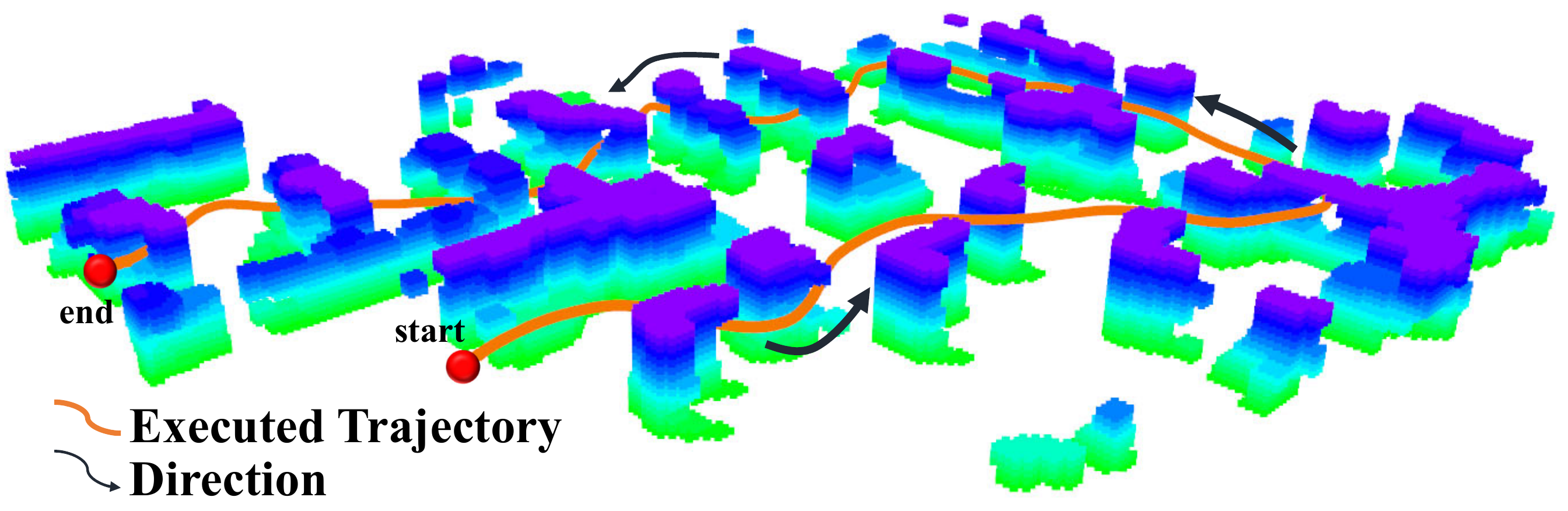}
	\vspace{-0.3cm}
	\captionsetup{font={small}}
	\caption{
		Trajectories executed by the quadrotor and online generated map which is colored by height in the system-level indoor experiment.
	}
	\vspace{-0.4cm}
	\label{fig:indoor_rviz}
\end{figure}

In the second half of the experiment, as Fig. \ref{fig:indoor_tpv}(b) shows, we set several 2m-width gates, 1m-diameter rings and 0.8m-diameter rings. Note that our quadrotor's diameter is 0.4m and Mavic Air 2's diameter is about 0.5m.
As Fig. \ref{fig:indoor_rviz} shows, our system can pass through various types of obstacles flexibly and fast.
However, Mavic Air 2 can only pass through 2m-width gates.
The result proves that our \delete{gaze-enhanced }system has more applicability and robustness for different environments.
\delete{We refer the readers to the attached video for more details.}

\subsubsection{Outdoor Flight Test}
\label{sec:ExpOutdoor}

Finally, to illustrate the reliability and potentiality of this work, we test in the large-scale outdoor environment, which is $35 \times 35 m$ in scale.
Volunteers are asked to go through several gates and rings in order.
The average time consumption and maximum speed are 49.5s and 1.96 m/s for the proposed system. 
For the consumer UAV, the respective data is 83s and 1.60 m/s. 
The attached video shows our method is more smooth and fast. 


\begin{figure}[t]
	\centering
	\includegraphics[width=0.96\linewidth]{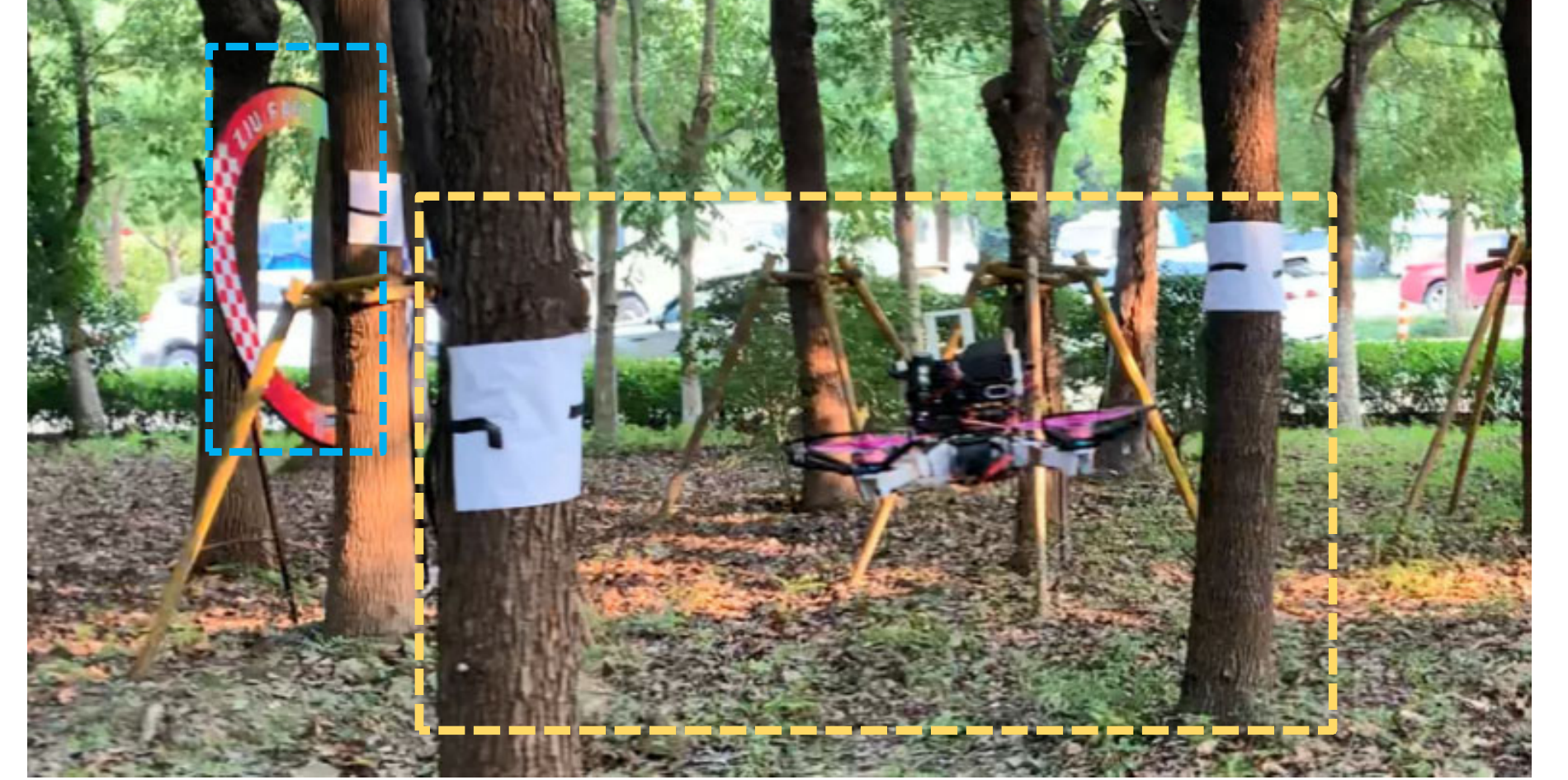}
	\vspace{-0.0cm}
	\captionsetup{font={small}}
	\caption{
		Flight in forest, the blue box marks a ring and yellow box marks a gate made from two trees pasted with white paper.
	}
	\vspace{-0.2cm}
\end{figure}
\begin{table}[ht]
	\centering
	\caption{Indoor System-level Comparison}
	\vspace{-0.1cm}
	\label{tab:IndoorSysComp}
	\begin{tabular}{@{}ccc@{}}
		\toprule
		System          & Avg Time (s) & Max speed (m/s) \\ \midrule
		Ours            & \textbf{7.5} & \textbf{1.97}   \\
		DJI Mavic Air 2 & 31.2         & 0.8             \\ \bottomrule
	\end{tabular}
	\vspace{-0.4cm}
\end{table}

\section{Conclusion}
\label{sec:conclusion}

In this paper, we investigate the high correlation between gaze and intent and propose a gaze-enhanced perception-aware assistive aerial teleoperation.
With the gaze-enhanced intent capture method, we can capture the intent more precisely.
Then we generate a perception-aware trajectory that accompanies the intent timely and safe.
Extensive indoor and outdoor experiments and benchmark comparisons validate that our method is robust and friendly to even unskilled users.

\delete{more precise intention can be obtained.
	Then we generate a topological guiding path following the intention for the back-end optimization.
	Finally a perception-aware spatial-temporal trajectory optimizer is proposed to generate a safe trajectory which simultaneously enhances the visibility to \add{the operator's interested} environment.}
\delete{
	In the future, our assistive teleoperation system will be extended to dynamic environments to have a wider range of applications.
}


\newlength{\bibitemsep}\setlength{\bibitemsep}{0.0\baselineskip}
\newlength{\bibparskip}\setlength{\bibparskip}{0pt}
\let\oldthebibliography\thebibliography
\renewcommand\thebibliography[1]{%
	\oldthebibliography{#1}%
	\setlength{\parskip}{\bibitemsep}%
	\setlength{\itemsep}{\bibparskip}%
}
\bibliography{main}
\end{document}